\documentclass{article}

\PassOptionsToPackage{numbers, compress}{natbib}

     \usepackage[final]{neurips_2022}

\DeclareUnicodeCharacter{0301}{\'{e}}
\usepackage[utf8]{inputenc} 
\usepackage[T1]{fontenc}    
\usepackage{hyperref}       
\usepackage{url}            
\usepackage{booktabs}       
\usepackage{amsfonts}       
\usepackage{amsmath}
\usepackage{stmaryrd}
\usepackage{nicefrac}       
\usepackage{microtype}      
\usepackage{graphicx}      
\usepackage{xcolor}         
\usepackage{enumitem}         
\usepackage{subcaption}         
\newtheorem{theorem}{Theorem}

\newcommand{\astar}{A*\ }
\newcommand{\logloss}{\ensuremath{\mathrm{L}_l}}
\newcommand{\loss}{\ensuremath{\mathrm{L}}}
\newcommand{\starloss}{\ensuremath{\mathrm{L}^*}}
\newcommand{\coatstar}{\ensuremath{\mathrm{CoAt}-\mathrm{L}^*}}

\title{A Differentiable Loss Function for Learning Heuristics in \astar}

\author{
  Leah Chrestien\\
  Department of Computer Science, FEL\\
  Czech Technical University in Prague\\
  \texttt{leah.chrestien@aic.fel.cvut.cz}
   \And
    Tomas Pevny \\
    Department of Computer Science, FEL \\
     Czech Technical University in Prague\\
   \texttt{pevnytom@fel.cvut.cz} \\
   \AND
   Antonin Komenda \\
    Department of Computer Science, FEL \\
     Czech Technical University in Prague\\
   \texttt{antonin.komenda@fel.cvut.cz} \\
   \And
   Stefan Edelkamp \\
   Department of Computer Science, FEL\\
    Czech Technical University in Prague\\
   \texttt{stefan.edelkamp@fel.cvut.cz} \\
}

\begin{document}

\maketitle
\begin{abstract}
  Optimization of heuristic functions for the \astar algorithm, realized by deep neural networks, is usually done by minimizing square root loss of estimate of the cost to goal values. This paper argues that this does not necessarily lead to a faster search of \astar algorithm since its execution relies on relative values instead of absolute ones. As a mitigation, we propose a \starloss{} loss, which upper-bounds the number of excessively expanded states inside the \astar search. The \starloss{} loss, when used in the optimization of state-of-the-art deep neural networks for automated planning in maze domains like Sokoban and maze with teleports, significantly improves the fraction of solved problems, the quality of founded plans, and reduces the number of expanded states to approximately 50\%.
\end{abstract}

\section{Introduction}
Automated planning aims to find a sequence of actions that will reach a goal in a model of the environment provided by the user. Planning is considered to be one of the core problems in Artificial intelligence and it is behind some of its successful applications~\cite{samuel1967some,knuth1975analysis,silver2017mastering}. Early analysis of planning tasks~\cite{McDermott96} indicated that optimising the heuristic function steering the search for a given problem domain can dramatically improve the performance of the search. Optimising heuristic function in fully automatic manner therefore becomes a central request in improving the performance of planners~\cite{bonet2001planning,thebook}.

Learning in planning means optimizing heuristic functions from plans of already solved problems and their instances. This definition includes selection of proper heuristics in a set of pattern databases~\cite{FrancoTLB17,HaslumBHBK07,MoraruEFM19,Edelkamp06}, a selection of a planner from a portfolio~\cite{katz2018delfi}, learning planning operators from instances~\cite{MenagerCRA18,Wang94}, and learning for macro-operators and entanglements~\cite{Chrpa10,korf1985macro}. Recent years observe a renewed interest in learning heuristic functions and this is fuelled by the success of deep learning and reinforcement learning in the same area~\cite{shen2020learning, groshev2017learning,ferber2020neural,bhardwaj2017learning}.

In this work, we are interested in optimising the heuristic function for \astar~\cite{hart1968formal}, which despite the popularity of Monte Carlo tree search~\cite{coulom2006efficient,silver2017mastering} is interesting due to its guarantees on optimal solution. \astar is also optimally efficient in the sense that it expands the minimal number of states. Majority of prior art~\cite{shen2020learning, toyer2020asnets, groshev2017learning,ferber2020neural,bhardwaj2017learning} optimises the heuristic function by minimizing the error of the predicted cost to the goal on a \emph{training set} of problem instances,\footnote{The training set contains solved problem instances, where the solution should be ideally found by a search finding optimal solution, such as \astar with ideally admissible heuristic function.} where the error is measured by the $\loss_2$ error function or its variant. The zero $\loss_2$ does not guarantee the optimal efficiency of \astar, hence it gives a false sense of security. Some of these deficiencies have already been pointed out in~\cite{wilt2016effective,ferber2020neural,frances2019generalized,vlastelica2021neuro}.

We propose a \starloss{} loss function tailored for \astar, which minimizes an upper bound on the number of expanded states. This is achieved by stimulating states on an optimal path to have a smaller cost function $f = g + h$ than those off the optimal path. By this, \starloss{} effectively utilizes all the states generated during the exploration of \astar, providing much more information to the learner. If \starloss{} on a given problem instance is equal to zero, it is guaranteed that \astar will expand only states on the optimal path, which under conditions on the training set as detailed below, implies optimal efficiency of \astar. We emphasize that the optimal efficiency is retained even on problems with exponentially many optimal paths~\cite{HelmertR08}, therefore the heuristic function has to learn a tie-breaking mechanism.

The proposed \starloss{} is compared to state of the art on the Sokoban problem and Maze with teleports, where it consistently outperforms heuristic functions optimizing $\loss_2.$ Sokoban was chosen because it is a PSpace complete problem, popular as a benchmark in many state of the art works. The experimental results shows that \astar with a heuristic function minimizing \starloss{} expands smaller number of states (as less as 50\%) and generalizes to bigger problems.

\section{Preliminaries}
We define a search problem instance by a directed weighted graph $\Gamma = \langle \mathcal{S}, \mathcal{E}, w \rangle$, a distinct node $s_0\in\mathcal{S}$ and a distinct set of nodes $\mathcal{S}^*\subseteq\mathcal{S}$. The nodes $\mathcal{S}$ denote all possible states $s\in\mathcal{S}$ of the underlying transition system representing the graph. The set of edges $\mathcal{E}$ contains all possible transitions $e\in\mathcal{E}$ between the states in the form $e = (s, s')$. $s_0 \in \mathcal{S}$ is the initial state of the problem instance and $\mathcal{S}^* \subseteq \mathcal{S}$ is a set of allowed goal states.  Problem instance graph weights (alias action costs) are mappings  $w : \mathcal{E} \rightarrow \mathbb{R}^{\ge 0}$.

Let $\pi=({e}_1, {e}_2, \ldots, {e}_l)$, we call $\pi$ a path (alias a plan) of length $l$ solving a task $\Gamma$ with $s_0$ and $\mathcal{S}^*$ iff $\pi=((s_0, s_1), (s_1, s_2), \ldots, (s_{l-1}, s_l))$ and $s_l \in \mathcal{S}^*$. An optimal path is defined as a minimal
cost of a problem instance $\Gamma, s_0, \mathcal{S}^*$ and is denoted as $\pi^*$ together with its value $f^*= w(\pi^*) = w({e}_1) +w({e}_2) + \ldots, + w({e}_l)$. We often minimize the
cost of solution of a problem instance $\Gamma, s_0, \mathcal{S}^*$, namely $\pi^*$, together with its length $l^*=|\pi^*|$.

\subsection{\astar algorithm}
Let's briefly recall how the \astar algorithm works. For consistent heuristics, where $h(s) - h(s') \leq w(s,s')$ for all edges $(s,s')$ in the $w$-weighted state space graph, it mimics the working of Dijkstra's shortest-path algorithm~\cite{Dijkstra59} and maintains the set of generated but not expanded nodes in $\mathcal{O}$ (the Open list) and the set of already expanded nodes in $\mathcal{C}$ (the Closed list). It works as follows.
\begin{enumerate}[itemsep=0mm]
	\item Add the start node $s_0$ to the Open list $\mathcal{O}_0$.
    \item Set $g(s_0) = 0$
	\item Initiate the Closed list to empty, i.e.  $\mathcal{C}_0 = \emptyset.$
	\item For $i \in 1,\ldots$ until $\mathcal{O}_i \neq \emptyset$
	\begin{enumerate}
		\item Select the state $s_i = \arg \min_{s\in \mathcal{O}_{i-1}} g(s) + h(s)$
		\item Remove $s_i$ from $\mathcal{O}_{i-1},$ $\mathcal{O}_{i} = \mathcal{O}_{i-1} \setminus \{s_i\}$
		\item If $s_i \in \mathcal{S}^*$, i.e. it is a goal state, go to 4.
		\item Insert the state $s_i$ to $\mathcal{C}_{i-1}$, $\mathcal{C}_i = \mathcal{C}_{i-1} \cup \{s_i\}$
		\item Expand the state $s_i$ into states $s'$ for which hold $(s_i, s') \in \mathcal{E}$ and for each
		\begin{enumerate}
		    \item set $g(s') = g(s_i) + w(s_i, s')$
		    \item if $s'$ is in the Closed list as $s_c$ and $g(s') < g(s_c)$ then $s_c$ is reopened (i.e., moved from the Closed to the Open list), else continue with (e)
		    \item if $s'$ is in the Open list as $s_o$ and $g(s') < g(s_o)$ then $s_o$ is updated (i.e., removed from the Open list and re-added in next step with updated $g(\cdot)$), else continue with (e)
		    \item add $s'$ into the Open list
		\end{enumerate}
	\end{enumerate}
	\item Walk back to retrieve the optimal path.
\end{enumerate}

In the above algorithm, $g(s)$ denotes a function assigning an accumulated cost $w$ for moving from the initial state ($s_0$) to a given state $s$.
Consistent heuristics are called monotone because the estimated cost of a partial solution $f(s)=g(s)+h(s)$ is monotonically non-decreasing along the best path to the goal.
More than this, $f$ is monotone on all
edges $(s,s')$, if and only if
$h$ is consistent as we have
$f(s') = g(s') + h(s') \ge g(s) + w(s,s') + h(s) - w(s,s') =
f(s)$ and $h(s) - h(s') = f(s) -g(s) - (f(s') -g(s'))
= f(s) - f(s') + w(s,s') \leq w(s,s')$.
For the case of consistent heuristics, no reopening (moving back nodes from Closed to Open) is needed, as we essentially traverse a state-space graph with edge weights $w(s,s') + h(s') - h(s) \ge 0$. For the trivial heuristic $h_0$, we have $h_0(s)=0$ and for perfect heuristic $h^*$, we have $f(s)=f^*=g(s)+h^*(s)$
for all nodes  $s$. Both heuristics $h_0$ and $h^*$ are consistent.

Even if the heuristic is not consistent, algorithms like \astar even without the reopening, remain complete i.e. they find a plan if there is one. Plans might not be provably optimal but are often very good in planning practice.

\subsection{Optimizing the heuristic}
We consider heuristic function $h_\theta:\mathcal{S} \rightarrow \mathbb{R}^{\ge 0}$ mapping a state $s\in\mathcal{S}$ to a real non-negative value, where $\theta \in \mathbb{R}^m$ holds parameters of $h_{\theta}$. Using a set of problem instances $\mathcal{T}$ (further called training set), we want to optimize parameters $\theta$ of $h_{\theta}$ such that an \astar search algorithm would find an (optimal) solution by expanding the least number of states. \footnote{Here it is assumed that the number of expanded states is directly proportional to the time taken to find the solution, as the time to compute the value of $h_\theta$ is independent of the value of $\theta$.} This, in practice, means to solve the optimization problem
\begin{equation}
\arg \min_{\theta} \sum_{\mathcal{S} \in \mathcal{T}} \loss(h_\theta, \mathcal{S}),
\label{eq:loss_on_set}
\end{equation}
where the loss function $\loss$ should be such that smaller values imply better heuristic function $h_\theta$ as perceived by \astar.
\subsection{\texorpdfstring{%
                Weakness of $\mathrm{L}_2$ loss function
                }{%
                Bookmark Version
            }}
\label{sub:l2loss}
Many prior art on optimizing heuristic function~\cite{shen2020learning,groshev2017learning,ferber2020neural,bhardwaj2017learning,toyer2020asnets}
minimize the $\loss_2$ loss function\footnote{While some works use $L_1$, the properties discussed here for $L_1$ holds for $\mathrm{L}_2$ as well.} $\loss_2(h_\theta, \mathcal{S}) = \sum_{i} \left(h_\theta(s_i) - y_i\right)^2,$ where the training set $\mathcal{S}$ consists of pairs $\left\{(s_i, y_i)\right\}_{i=1}^n,$ where $s_i$ is some state and $y_i$ is the length of the plan from $s_i$ to the goal state. We argue that zero $\loss_2$ loss on a given \emph{problem instance} for states on the optimal path does not guarantee that \emph{\astar will be optimally efficient} in the sense that it can expand more states than needed.

\begin{figure}[t]
\begin{center}
\centering
  \begin{subfigure}[t]{1.5in}
    \centering
    \includegraphics[width=1.5in]{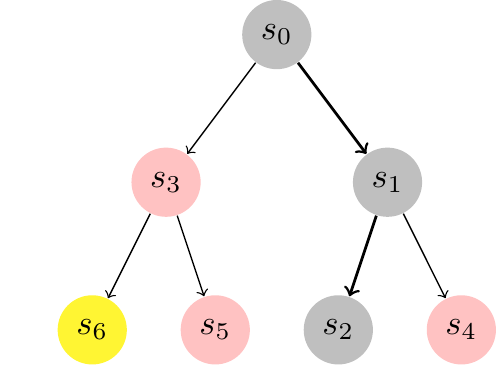}
    \caption{}\label{fig:astara}
  \end{subfigure}
  \hspace{1in}
  \begin{subfigure}[t]{1.5in}
    \centering
    \includegraphics[width=1.5in]{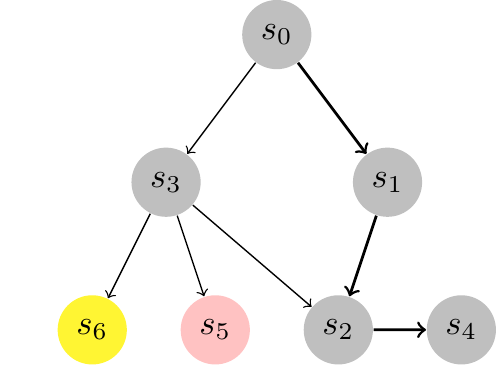}
    \caption{}\label{fig:astarb}
  \end{subfigure}
\caption{\label{fig:astar}A visualization of a search space of an \astar algorithm. In sub-figure (a), path $s_0 \rightarrow s_1 \rightarrow s_2$ represents the optimal plan, states $\{s_3, s_4, s_5\} / s_6$ are off the optimal path but have / have not been generated by the \astar. In sub-figure (b), path $s_0 \rightarrow s_1 \rightarrow s_2 \rightarrow s_4$ and $s_0 \rightarrow s_1 \rightarrow s_3 \rightarrow s_4$ represents the optimal plan, states $s_5 / s_6$ are off the optimal path but have / have not been generated.}
\end{center}
\end{figure}
\textbf{L2 does not utilize states off the optimal path.}  Imagine a problem instance shown in Figure~\ref{fig:astara}, where $s_0$ and $s_2$ are the initial and goal states respectively and $(s_0, s_1, s_2)$ is the optimal path. When L2 loss is optimized, one needs to know the exact cost-to-go values which are obtained by solving the problem instance. Thus, by solving the instance in Figure~\ref{fig:astar}, one obtains heuristic values for $\{s_0, s_1, s_2\}.$ But if $\loss_2$ loss is equal to zero on them, it does not say anything about estimates for states off the optimal path states $\{s_3, s_4, s_5, s_6\}.$ This means that it can happen that $f(s_3) < f(s_1),$ which would lead to expanding the state $s_3$ in \astar algorithm and hence to sub-optimality.

This issue can be fixed, if the training set is extended to contain heuristic values for all states off the optimal path ($\{s_3,s_4,\ldots,s_6\}$ in our example), which in practice requires solving all possible variants of the problem instances. This has been suggested in~\cite{bhardwaj2017learning} but is infeasible due to excessive computational requirements. Therefore, in practice, it is assumed the training set to be large, thereby mitigating this problem.

\textbf{$\loss_2$ loss provides a false sense of optimality.} Some problems can have large (even exponential) number of optimal solutions with the same cost~\cite{HelmertR08}. In this case, minimizing estimate of cost-to-go of all states in the problem instance (to fix the problem mentioned above) does not guarantee that \astar will be optimally efficient. Consider an example in Figure~\ref{fig:astarb} with unit cost on edges. The algorithm starts by expanding state $s_0$ to $s_1$ and $s_2$. The heuristic is perfectly estimated and $f(s_1) = f(s_2) = 3.$ Hence there are two states with the same value $f$ which means \astar has to decide, how to the break this tie. The situation repeats after \astar expands either $s_1$ or $s_3,$ since the open set will now contain state $s_3$ with $f(s_3) = 3$ and \astar needs to resolve ties again. See~\cite{HelmertR08} for more examples.

\textbf{Heuristic value for unreachable (dead-end) states} should be infinite to ensure that they are never selected. An infinity in the $\loss_2$ loss would always lead to an infinite loss which would then result in an infinite gradient. Hence, in practice, a sufficiently large value for dead-end states has to be used.
\section{\texorpdfstring{%
                \starloss{} loss
                }{%
                Bookmark Version
            }}
We explain the proposed \starloss loss function on a single problem instance $\Gamma = \langle \mathcal{S}, \mathcal{E}, w\rangle $ (the extension to a set of plan is trivial through Equation~\eqref{eq:loss_on_set}).  We assume to have a (preferably optimal and shortest) plan $\pi = ((s_0, s_1), (s_1, s_2), \ldots, (s_{n-1}, s_n))$ with states from this optimal plan denoted as $\mathcal{S}^o= \{s_0, s_1, s_2, \ldots, s_n\}.$ This plan can be found by \astar with some (admissible) heuristic function $h$, which \emph{does not have to} coincide with the heuristic function $h_\theta$ that we are optimizing. We denote states off the optimal plan as $\mathcal{S}^n \subset \mathcal{S} \setminus \mathcal{S}^o,$ where the subset exists because, in practice, $\mathcal{S}^n$ contains states generated by \astar while solving the problem instance $\Gamma.$ In the visualization in Figure~\ref{fig:astara}, grey states are on the optimal path $\mathcal{S}^{o},$ pink states are off the optimal path, and yellow states were not generated in the course of solving the problem instance. Hence, $\mathcal{S}=\{s_i\}_{i=1}^6,$ $\mathcal{S}^o=\{s_i\}_{i=1}^2$ and $\mathcal{S}^n=\{s_i\}_{i=3}^5.$ The training sample for a \starloss is defined as a tuple $\bar{\Gamma} = \langle \mathcal{S}, \mathcal{E}, w\rangle, \mathcal{S}^o.$

\starloss{} aims to minimize the number of expanded states in the \astar algorithm. Recall that \astar always expands a state from an open list with smallest $f_{\theta}(s) = g(s) + h_{\theta}(s).$ To be optimally efficient, states on the optimal path $s' \in \mathcal{S}^{o}$ should have \emph{always} smaller $f(s')$ than states off the optimal path $s'' \in \mathcal{S}^{n},$ i.e.
\begin{equation}
(\forall s' \in \mathcal{S}^o)(\forall s'' \in \mathcal{S}^n)(g(s') + h_{\theta}(s') < g(s'') + h_{\theta}(s''))
\label{eq:constraints1}
\end{equation}
On the optimal path, we might also impose monotonicity as
\begin{equation}
(\forall s_i, s_j \in \mathcal{S}^o)(i<j)(g(s_i) + h_{\theta}(s_i) \leq g(s_j) + h_{\theta}(s_j),
\label{eq:constraints2}
\end{equation}
though it does not affect the optimality of \astar. We do this, since monotonic heuristic function implies \astar returning optimal solution. In Constraint~\eqref{eq:constraints1}, states not generated by \astar are ignored. But $\mathcal{S}^n$ will always contain all states of distance one from the optimal path, which is sufficient to show that a loss equalling zero implies expanding states only on an optimal path (in the training set). To prevent confusion, we emphasize that conditions are designed for the heuristic $h_{\theta}$ that is to be optimized, and not for the heuristic $h$ that has generated the training set in the first place.

While Constraint~\eqref{eq:constraints2} is true for every consistent heuristic, Constraint~\eqref{eq:constraints1} is true only for perfect heuristics. Otherwise, we could have some
earlier states in the exploration off the optimal path that have a smaller $f$-value than later ones in the optimal path.
What seems to be over-restrictive, such that almost no heuristic function will ever fulfill, Constraint ~\eqref{eq:constraints1}, is intentional. 

The proposed $\starloss$ loss minimizes the number of times each of the above constraints are violated as
\begin{alignat}{2}
& \frac{1}{|\mathcal{S}^o||\mathcal{S}^n|}\sum_{s'\in \mathcal{S}^o}\sum_{s''\in \mathcal{S}^n} \llbracket g(s') + h_{\theta}(s') \geq g(s'') + h_{\theta}(s'')\rrbracket + \nonumber
\\
& \frac{1}{|\mathcal{S}^o|(|\mathcal{S}^o| - 1)}\sum_{i =2}^{|\mathcal{S}^o|}\sum_{j = 1}^{i}\llbracket g(s_i) + h_{\theta}(s_i) > g(s_j) + h_{\theta}(s_j)\rrbracket, \label{eq:loss}
\end{alignat}
where $\llbracket\cdot\rrbracket$ is an Iverson bracket, which is equal to one if the argument is true and zero otherwise. The first part of the loss function loosely upper bounds the number of non-optimal states the \astar expands while the second part ensures the monotonicity of the heuristic function along the optimal plan.
In other words, the conditions~\eqref{eq:constraints1} and~\eqref{eq:constraints2} encode the aim of a consistent and perfect heuristic. During training, we iterate over many samples of \astar explorations which enlarges the scope of $\starloss.$

For setting up constraints for heuristic learning, we only need the partitioning of the set of explored nodes into the sets
$S^0$ and $S^n$, computed via an optimal plan and
a set of all generated nodes, together with their $g$-values.
Given the optimal heuristic, \astar will always find an optimal solution. Up to tie-breaking, it is optimally efficient and will expand only nodes with optimal merit $f^*$.

Loss function $\starloss$ does not distinguish between the Open and Closed lists in the exploration of \astar as long as it has access to the combined set of explored nodes. This way, we can take any optimal planner and not just the heuristic search planners for training.
\subsection{\texorpdfstring{
                }{%
                Bookmark Version
            }}
\textbf{$\starloss$ utilizes all states} generated during the \astar search used to create the training sample(s), which is in sharp contrast to $\loss_2$ estimating cost-to-go. This propagates to better utilization of states in the training set. The experimental results show that given a fixed and small number of training problems, models minimizing $\starloss$ achieves higher performance.

\textbf{$\starloss=0$ implies optimality.}
We state a following theorem:
\begin{theorem}[Upper Bound]
For a problem instance with states $\mathcal{S} =\mathcal{S}^n \cup \mathcal{S}^o,$ denote
\begin{equation}
\mathcal{R}^n =
\left\{s'' \in \mathcal{S}^n \mid \exists s' \in \mathcal{S}^0  \wedge \left( g(s') + h_{\theta}(s') \geq  g(s'') + h_{\theta}(s'') \right) \right\},
\end{equation}
the quantity $|\mathcal{R}^n|$  is an upper bound on the number of non-optimal states \astar expands during its search.
\end{theorem}
The proof is straightforward and it is included in supplementary for completeness. The quantity $|\mathcal{R}^n|$ is exactly the quantity minimized by the $\starloss$ as defined in Equation~\eqref{eq:loss}. The following theorem is a trivial consequence of this property.
\begin{theorem}[Optimal efficiency]
Let for a given training sample $\bar{\Gamma},$ and a heuristic function $h_\theta$ $\starloss(\bar{\Gamma, h_\theta}) = 0.$ Then \astar{} with heuristic function $h_\theta$ will expand only states on the optimal path $\mathcal{S}^o.$ If $\mathcal{S}^o$ in $\bar{\Gamma}$ is optimal and shortest, \astar{} will be optimally efficient.
\end{theorem}
The proof is a consequence of the property that $\starloss(\bar{\Gamma, h_\theta}) = 0$ implies that $|\mathcal{R}^n| = 0.$
The above theorem holds even on problems with multiple optimal solutions. In this case, \starloss{} would be either equal to zero, which means $h_\theta$ includes tie-breaking mechanism and it will be optimal, or it will be greater than zero. Thus and unlike $\loss_2,$ its zero value implies optimal efficiency.

\textbf{\starloss does not require heuristic value of unreachable (dead-end) states,} which is caused by the fact that \starloss requires satisfaction of inequalities instead of estimation of some value.
\starloss{} loss does not force the heuristic to be goal aware, since as discussed in Supplementary, this is not needed for the optimal efficiency of \astar.

\section{Related Work}
In potential heuristics~\cite{seipp2015new}, parameters of the heuristic functions are optimized by linear programming for a particular problem instance to satisfy constraints similar to those stated in this paper. The optimization assumes a particular structure of the heuristic, unlike here, where no structure is assumed. Ref.~\cite{takahashi2019learning} admits that the symmetry of $\loss_2$ (and of $\loss_1$) loss does not promote admissibility of the heuristic. It recommends asymmetric $\loss_1$ with different weights on the left and right parts, but this does not completely solve the problems identified above in Section~\ref{sub:l2loss}.

Ref.~\cite{wilt2016effective} recognizes that magnitude of heuristic is not important and suggest to measure the quality with a correlation to a distance to goal, which neither solves the problem with ties nor does it utilize states off optimal path. In~\cite{ferber2020neural}, neural networks estimate the number of expansions of a GBFS search, though the results are comparable to an estimation of cost-to-goal. In~\cite{bhardwaj2017learning} \astar is viewed as a Markov Decision Process with value function being equal to the number of steps of \astar till it reaches the solution. While this detaches the heuristic values from cost-to-goal cost, it does not solve the problem with dead-ends, state efficiency, and ties.

Refs.~\cite{vlastelica2021neuro,yonetani2021path} combine neural networks with discrete search algorithms, which become an inseparable part of the architecture. 
Our setting is more classical where the heuristic is optimized for \astar search but the execution of the search is independent of the training. This has the advantage that one (costly) execution of \astar search algorithm is used many times during training to optimize weights.

A large corpus of literature~\cite{silver2017mastering,guez2018learning,feng2020solving,anthony2017thinking} is devoted to improvements to Monte Carlo Tree Search. Since this work is concerned exclusively to \astar algorithm, we view these works independent to this. Nevertheless, we compare to some of them in the experimental section. Similarly, a  lot of works~\cite{shen2020learning, toyer2020asnets,zhang2020extending, groshev2017learning,chrestian2021,ferber2020neural,bhardwaj2017learning} investigate architectures of neural networks for learning a heuristic function, ideally for arbitrary planning problems. These works are perpendicular to this one, which investigate how to optimize these neural networks to perform well inside \astar.

\section{Experimental evaluation}
Heuristic functions optimized with respect to $\starloss$ loss function are compared to those optimized with respect to $\loss_2$ loss on two domains: Sokoban and Mazes with teleports. This is supplemented by the comparison with domain-independent planners: (1) SymB\astar \cite{torralba2014symba}, a cost-optimal planner from International Planning Competition (IPC) 2014; (2) Delfi~\cite{katz2018delfi}; (3) Mercury14 \cite{katz2014mercury}, a satisfycing planner from IPC 2014; (4) Stone soupe~\cite{seipp2018fast}, and by solutions based on Monte Carlo Tree Search~\cite{guez2018learning} and reinforcement learning~\cite{racaniere2017imagination,guez2019investigation}. Since authors of~\cite{toyer2020asnets} admit that their solution does not work on Sokoban problems, and~\cite{shen2020learning} works only on small Sokoban problems with two boxes, we do not compare to these works. All experiments involving neural networks have been repeated three times.

\paragraph{The Neural Network} Neural networks (NN) implementing heuristic functions (see Appendix for the full structure) were adopted from~\cite{groshev2017learning} and~\cite{chrestian2021}, where the latter is, to our best knowledge the state-of-the-art architecture for maze domains. It contains seven convolution layers $P_1, \ldots P_7$ followed by four convolution-attention-position blocks, which allow correlating information from distant parts of the maze. The output tensor of the fourth CoAt block is "flattened" by global average pooling over the $x$ and $y$ dimension to a vector, which is then fed to a fully connected layer ($\mathrm{FC}$) which outputs a scalar estimating the heuristic value. For more details on the network, we refer the reader to the original publication~\cite{chrestian2021}. This network is further called "CoAt" after the CoAt blocks. We also studied the network of~\cite{groshev2017learning} (without policy head), which has a similar structure, but instead of four CoAt blocks it has seven CNN layers. We refer to this network as to CNN. Both networks are by design scale-free, which means that they can be used on mazes of various sizes, as is shown below on the mazes with teleport domain. Since the \starloss loss as defined in Equation~\ref{eq:loss} is not differentiable, the Iverson bracket is replaced by a logistic loss function $\logloss(x) = \log(1 + \exp(-x)).$

Our experiments were implemented in Keras-2.4.3 with Tensorflow-2.3.1 as the backend. For training the neural networks, we used an NVIDIA Tesla GPU model V100-SXM2-32GB; the evaluation was performed on the CPU to ensure a fair comparison to domain independent planners. The neural networks were trained by the Adam optimizer \cite{kingma2014adam} with a default learning rate of 0.001. Each mini-batch contained \textbf{all} states from one problem instance. Scripts reproducing our experiments together with mazes and solutions will be made available upon acceptance.

\subsection{Training from solved mazes}
We generate mazes for the training set by running the \astar algorithm using the heuristic function from~\cite{groshev2017learning} on a set of problem instances to identify a set of states generated during the \astar search. All these sets form the training set. 
Since optimizing the heuristic by $\loss_2$ loss requires knowing the true heuristic value (cost to reach the goal), we have used SymB\astar \cite{torralba2014symba} to find the optimal plan from each state in the training set. For states for which SymB\astar doesn't find a solution (dead-end states), the \textit{h} value is replaced by a very large value. This construction, albeit very expensive, allows a fair comparison, since the training of heuristic by $\loss_2$ loss will also use states off the optimal path of the original problem instance for which the states were generated.

\textbf{Sokoban}'s training set contained 10000 Sokoban mazes of size $10 \times 10$ with 3 boxes created using gym-sokoban \cite{SchraderSokoban2018}. The testing set contained 2000 mazes of the same size $10 \times 10$ but with $3, 4, 5, 6, 7, 8, 9$ boxes. The complexity increases with the addition of more boxes.\footnote{This is of course just an approximation, as we can have simple problems with a large number of boxes}, and therefore we can evaluate the ability to generalize outside training environments. We go a step further and implement curriculum learning \cite{bengio2009curriculum} by training from those mazes that are solved by our network during evaluation. We create a new training set containing all the solved mazes and re-train our network in an effort to improve the coverage of our network.

\textbf{Maze-with-teleports}'s training set contained 5000 randomly generated mazes of size $n \times n = 15 \times 15$ with the agent in the upper-left corner and the goal in the lower-right corner. The mazes were generated using an open-source maze generator \footnote{https://github.com/ravenkls/Maze-Generator-and-Solver}, where walls were randomly broken and 4 pairs of teleports were randomly added to the maze structure. The testing contained 2000 mazes generated by the same algorithm but (i) were bigger by up to $60 \times 60$ and (ii) were rotated by 90, 180, and 270 degrees which moved the start and goal states to positions not occurring in the training set.
\begin{table*}[t]
\centering
\begin{subtable}[b]{3in}
\setlength{\tabcolsep}{3pt}
\small{
\begin{tabular}{c| c c c c c c c c c c c c}
\multicolumn{2}{c}{} & & & &\multicolumn{2}{c}{CNN} & &\multicolumn{3}{c}{CoAt}\\
\cline{6-7}
\cline{9-11}
$\#b$ &          SB\astar & Delfi & Merc & FDSS & $\loss_2$ & $\starloss$ & & $\loss_2$ & $\starloss$ & CL w.$\starloss$ \\
\hline
3 & 100 & 100 & 100 & 100 & 81 & 87   & &91 & 94 & 95\\
4 & 100 & 100 & 81 &100  &74& 80 & &89 & 93  & 94\\
5 & 97 & 91 & 67 & 94 &72 &82 & & 85 & 89  & 90\\
6 & 55 & 55 & 49 & 56 & 61 &71 & & 73 & 80   & 85\\
7 & 46 & 44 & 31 & 42 & 51 &59 & & 63 & 77  & 83\\
8 & - & - & - & - & - & - & & - & 32  & 59\\
9 & - & - & - & - & - & - & & - & 12  & 38\\
\hline
\end{tabular}
}
\caption{}
\label{tab:percentcov}
\end{subtable}
\hspace{0.2in}
\begin{subtable}[b]{2in}
\centering
\setlength{\tabcolsep}{4pt}
\small{
\begin{tabular}{lc}
model & coverage \\ \hline
MCTSNet & 84 \\
I2A & 84 \\
DRC (3,3) 10k & 93 \\
DRC (3,3) 900k & 99 \\
\coatstar & 97 \\
CL \coatstar & 100 \\ \hline
SBA* & 100 \\ \hline
\end{tabular}}
\caption{}
\label{tab:boxobancoverage}
\end{subtable}

\caption{\textbf{Left:} Fraction of solved mazes (in percents) of S(ym)B\astar, Delfi(1), Merc(ury14), FDSS (Fast Downward Stone Soup), CoAt and CoAt* on test data sets containing variable number of boxes. Column captioned $\#b$ indicates the number of boxes in different categories. The standard deviation of all repeated experiments was between $0.004$ and $0.008$ and it is not shown to save space.
\textbf{Right:} Fraction of solved mazes (in percents) from Boxoban dataset. DRC 900k / DRC 10k optimizes on $10k$ levels. Results of MCTSNet~\cite{guez2018learning}, I2A~\cite{racaniere2017imagination}, and DRC~\cite{guez2019investigation} have been copied from Table 2 of~ \cite{guez2019investigation}.}
\end{table*}

\subsection{Results}
\begin{figure}[t]
\centering
\begin{subfigure}[t]{2.1in}
  \centering
  \includegraphics[width=2.1in]{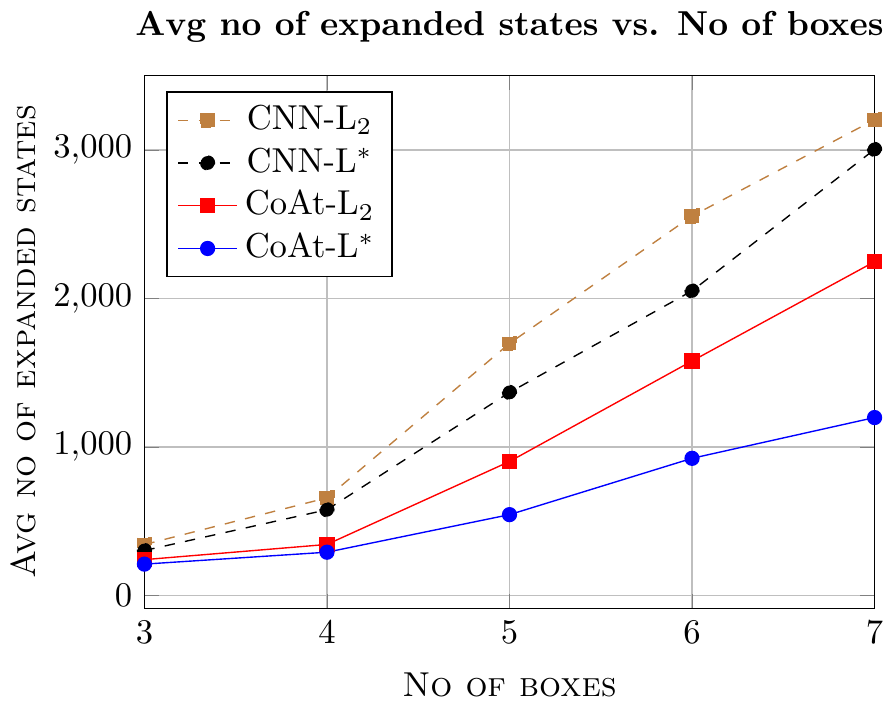}
  \caption{Sokoban}\label{fig:exp_states_maze}
\end{subfigure}
\hspace{0.5in}
\begin{subfigure}[t]{2.1in}
  \centering
  \includegraphics[width=2.1in]{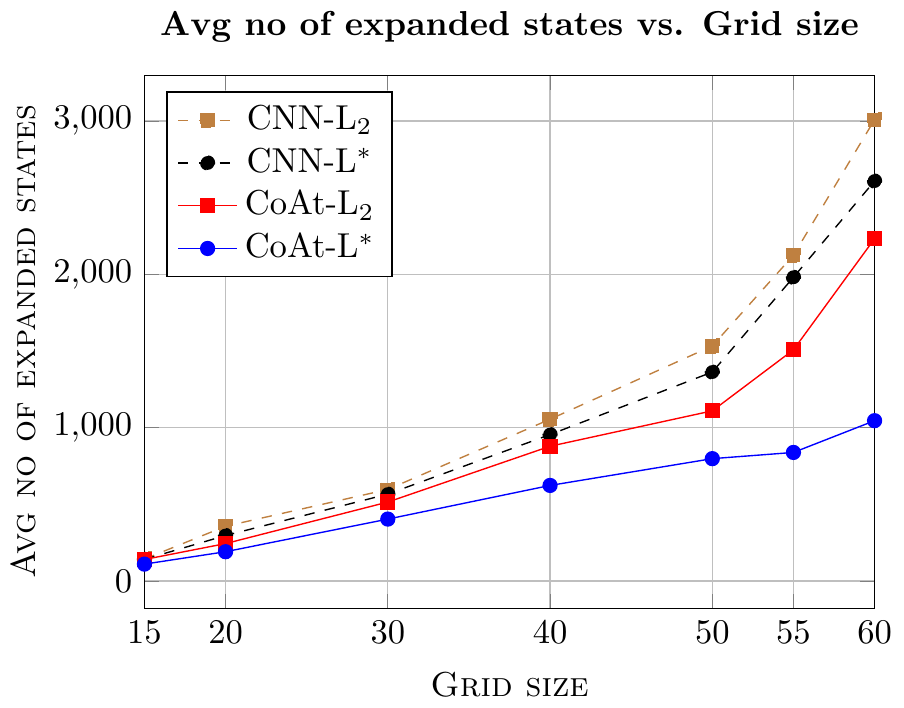}
  \caption{Maze with teleports}\label{fig:exp_states_sok}
\end{subfigure}
\caption{The average number of expanded states in \astar for Sokoban (left) and Maze problems (right).}
\end{figure}

\paragraph{Sokoban} Table \ref{tab:percentcov} shows the percentage of solved mazes of all compared planners on problem instances with a various number of boxes (recall that the NNs were optimized on instances with only three boxes). All planners were given a time limit of 10 minutes to solve each Sokoban instance. On mazes with 3 and 4 boxes, the optimal planners SymB\astar (SB\astar) and Delfi were able to solve all problem instances while the best performing architecture among the NNs, which is CoAt optimized with respect to $\starloss$ (\coatstar), could solve 94\% and 93\% of the mazes respectively. On increasing the number of boxes, the \astar with NNs start outperforming classical planners. \astar with NNs optimizing $\starloss$ is consistently better than those optimizing L$_2.$ The CoAt architecture proposed in~\cite{chrestian2021} with the proposed $\starloss$ is the only solver that can solve some mazes with 8 and 9 boxes. Solutions found by CoAt optimizing \starloss were close to optimum, on average by 1 step longer, which is likely because the learnt heuristic is most of the times admissible (see Supplementary for details). The average number of expanded states in Figure~\ref{fig:exp_states_maze} shows that NNs optimizing \starloss indeed expand a smaller number of states than those optimizing $\loss_2$.

Since networks have never seen mazes with four or more boxes, extrapolation to eight and nine boxes is impressive. To evaluate the potential for self-improvement, the training set of \coatstar was extended with mazes from the testing set it has already solved for fine-tuning. We refer to this as curriculum learning (CL w. \starloss) and the results are shown in the last column of Table \ref{tab:percentcov}. It shows marginal improvement on mazes with 3-5 boxes but records a significant improvement in performance over the vanilla \starloss on mazes exceeding 5 boxes.

\paragraph{Boxoban} On unfiltered "Boxoban" levels from~\cite{guez2019investigation}, \astar with heuristic implemented by CoAt network and optimized with respect to the proposed \starloss is compared to MCTSNet~\cite{guez2018learning}, Imagination augmented agent (I2A)~\cite{racaniere2017imagination}, and to DRC (3,3) network~\cite{guez2019investigation}, with possible  discrepancies, as results on the competing methods were taken from~\cite{guez2019investigation}. \coatstar was optimized on 10k mazes with 3 boxes (it is the same network as reported in the previous paragraph), whereas others were optimized on 900k mazes with 4 boxes. \coatstar and MCTSNet knows the model, whereas DRC and I2A do not. The fraction of solved mazes shown in Table~\ref{tab:boxobancoverage} shows that \coatstar trained on 10k mazes is second best behind DRC (3,3) that is trained on 900k mazes, but DRC (3,3) trained on 10k mazes is already inferior. \coatstar with one iteration of curriculum learning where the training set is extended to contain previously solved mazes (from set used in previous paragraph) solves 100\% of boxoban mazes. Needless to say that during optimization, DRC (3,3) allowed 1G iterations of SGD, MCTSNets allowed 10M iterations of SGD, whereas CoAt optimized \starloss allowed just for 120k iterations, which is several magnitudes less.

\begin{table*}[t]
\begin{center}
\setlength{\tabcolsep}{2pt}
\small{
\begin{tabular}[t]{c| c c c c c c c c c}
\multicolumn{3}{c}{} & & &\multicolumn{2}{c}{CNN} & &\multicolumn{2}{c}{CoAt}\\
\cline{6-7}
\cline{9-10}
$n$ & SB\astar & Delfi & Merc & FDSS & $\loss_2$ & $\starloss$ & & $\loss_2$ & $\starloss$ \\
\hline
$50$ & 92 & 90 & 75 & 100&100 & 100 & &100 &  100 \\
$55$ & 52 & 50 & - & 100&85 & 85 & &86 & 88\\
$60$ &  -   & - & - & 100&73 &74 & &76   & 79\\
\hline
\end{tabular}}
\caption{Fraction of solved mazes with teleports of S(ym)B\astar, Delfi(1), Merc(ury14), and \astar algorithm with heuristic function implemented by CNN and CoAt networks optimized with respect to $\loss_2$ and \starloss. All solvers have solved all mazes of size 15--40, hence they are not shown Standard deviations of repeated experiments were between 0.005 and 0.008 are shown in Suplementary.
}
\label{tab:mazecoverage}
\end{center}
\end{table*}

\textbf{Maze-with-teleports}
Fraction of solved mazes with teleports is shown in Table \ref{tab:mazecoverage}. \astar with heuristic implemented by NN was optimized on mazes of size $15 \times 15$ and has solved all mazes up to size $40 \times 40$ (not shown in the table) and beyond. The results mimic the results on Sokoban in the sense that \astar with CoAt networks optimizing $\starloss$ is consistently outperforming those optimizing $\loss_2.$

All 2000 mazes in the training set were created such that the agent starts in the top left corner and the goal is in the bottom right corner. When mazes are rotated by 90$^{\circ}$, 180$^{\circ}$ and 270$^{\circ},$ the agent has to solve mazes with distributions very different to that on the training set, yet the fraction of solved mazes for \coatstar decreases by at most 5\% (see Table 3 in Supplementary). Average number of generated states in \astar with different heuristics is shown in Figure \ref{fig:exp_states_maze}. Again, heuristics optimized with respect to \starloss expand smaller number of states during the search.

\begin{table*}[t]
\centering
\begin{subtable}[b]{3in}
\setlength{\tabcolsep}{3pt}
\small{
\begin{tabular}[b]{c| c c c c c c c c c c c c c}
\multicolumn{1}{l}{}  & &\multicolumn{2}{c}{3000} & &\multicolumn{2}{c}{4000} & &\multicolumn{2}{c}{5000}  & &\multicolumn{2}{c}{6000}\\
\cline{3-4}
\cline{6-7}
\cline{9-10}
\cline{12-13}
epoch & &$\loss_2$ & $\starloss$ & & $\loss_2$ & $\starloss$ & & $\loss_2$ & $\starloss$ &&$\loss_2$ & $\starloss$\\
\hline
0 && 7.4 & 7.4 && 7.7 & 7.7 && 8.5 & 8.5&  & 8.3 & 8.3 \\
1 && 11 & 11 && 9.8& 11 &&  10 & 10 && 9.6 & 10 \\
2 && 15 & 18 && 11 & 16 && 12 & 18 && 14 & 16 \\
3 && 29 & 33 && 20 & 31 && 21 & 24 && 20 & 32\\
4 && 34 & 62 && 27 & 69 && 36 & 49 && 31 & 48 \\
5 && 62 & 75 && 44 & 75 && 57 & 73 && 34 & 66 \\
6 && 67 & 77 && 51 & 80 && 61 & 80 && 49 & 77 \\
7 && 63 & 76 && 62 & 80 && 74 & 82 && 50 & 86 \\
\hline
\end{tabular}
}
\caption{Sokoban}
\label{tab:sokobanbootstrap}
\end{subtable}
\hspace{0.2in}
\begin{subtable}[b]{2in}
\centering
\setlength{\tabcolsep}{3pt}
\small{
\begin{tabular}[b]{c| c c c c c c c c c c c }
\multicolumn{1}{l}{}  & &\multicolumn{2}{c}{1000} & &\multicolumn{2}{c}{3000} & &\multicolumn{2}{c}{5000}\\
\cline{3-4}
\cline{6-7}
\cline{9-10}
epoch & &$\loss_2$ & $\starloss$ & & $\loss_2$ & $\starloss$ & & $\loss_2$ & $\starloss$\\
\hline
0 && 25 & 25 && 19 & 19 && 15 & 15 \\
1 && 42 & 45 && 29 & 35 && 18 & 20 \\
2 && 45 & 68 && 39 & 51 && 34 & 34 \\
3 && 69 & 83 && 59 & 67 && 41 & 59 \\
4 && 84 & 90 && 78 & 83 && 66 & 75\\
\hline
\end{tabular}
}
\caption{Maze with teleports}
\label{tab:mazebootstrap}
\end{subtable}

\caption{Fraction of solved mazes (in percents) when the networks are optimized only on mazes they have previously solved. First row corresponds to \astar with untrained network.}
\label{tab:bootstrap}
\end{table*}

\subsection{Training from unsolved mazes}
The training set for the above experiments required a number of solved mazes, which is expensive. Are they needed? Consider a following protocol, where in each iteration, a heuristic function implemented by the NN is first used in \astar to try to solve mazes from an available set of mazes (recall that we set a 10min time limit for solving a maze) and then to optimize its parameters on a set of mazes it has solved. Similarly to reinforcement learning, if an un-optimized (which means uninformed) heuristics solves at least a few mazes, it can jump-start the learning.

In this experiment, the training set of unsolved mazes is fixed. In the optimization of our NN over solved mazes, we perform one epoch. Hence the number of iteration and epoch coincide. Table \ref{tab:bootstrap} shows percentages of solved mazes on Sokoban and Maze with teleports for the first seven and four epochs (epoch number 0 means that the network is untrained) for different sizes of the training set. The set of Sokoban mazes contained problems with three, four, and five boxes; the set of mazes with teleports contained problems of size $40 \times 40$.

We observe that the fraction of solved mazes increases with epochs and the speed of this growth is significantly faster for heuristics optimized with respect to the proposed \starloss. To our surprise, the fraction of solved mazes does not grow faster when the number of initially unsolved set of mazes is bigger. Yet we have observed that the fraction of solved unfiltered boxoban mazes increases as expected. \astar with the network optimized on the set of 6000 mazes could solve 96\% of levels of unfiltered boxoban mazes after seven epochs. This agent has performed just 20.5k gradient descend steps, which is comparatively smaller than 1G steps of DRC (3,3) agent from~\cite{guez2018learning}.

\section{Conclusion}
This work has proposed \starloss{} loss function for imitation learning in planning;it has been designed specifically to maximize the efficiency of the \astar algorithm.  \starloss{} is zero, if and only if the basic monotonicity requirements on the f-value in \astar are satisfied, so that the heuristic trained on this loss function is consistent and perfect. This enables \astar to find the optimal solution at an optimal time. It has been shown that $\loss_2$ does not have these guarantees.
The experiments have verified the promises that \astar with heuristic functions optimized with respect to \starloss{} \textbf{always} solve a much higher number of problems, generate up to 50\% lesser states than those optimized with respect to the usual $\loss_2,$ and return nearly optimal solutions. By comparison to MCTSNets, we have shown that \astar with well trained heuristic can be competitive to Monte Carlo Tree Search. The training is also more efficient, as we use a much lesser number of SGD steps.

The proposed \starloss loss well complements contemporary research, which pays a lot of attention to the network architectures, as it can be used as a drop-in replacement for $\loss_2.$ It inspires us to design loss functions for other types of search algorithm and research neglected aspects, such as the construction of a representative training set. We see them as a limiting factor in the further endeavour to solve more difficult problem instances.

\section*{Statement of broader impact}
\astar is a workhorse optimization algorithm in industry owing its popularity to its simplicity and performance guarantees. Optimizing the heuristic function for a particular problem in order to maximize its efficiency is important for the following reasons: (i) to solve bigger problems; (ii) to find better solutions;  (iii) or to solve existing problems more efficiently thus saving costs / energy / CO2.
The important feature of our proposed loss function is its higher sample efficiency. Comparing to the prior art, it  requires smaller training sets and also leads to a faster convergence as lesser number of stochastic gradient descend steps (by a few orders of magnitude) are needed. Higher efficiency not only saves energy and CO2, but decreases the inequality in the research, as not all teams can afford to optimize large models over long period of time.

\bibliography{main.bib}

\end{document}